\pdfoutput=1

\documentclass[11pt]{article}

\usepackage[acceptedWithA]{tacl2021v1}

\usepackage{times}
\usepackage{latexsym}
\usepackage{graphicx} 
\usepackage{enumitem}
\usepackage[T1]{fontenc}

\usepackage[utf8]{inputenc}

\usepackage{microtype}
\usepackage{inconsolata}
\usepackage{booktabs}
\usepackage{adjustbox}
\usepackage{subcaption}
\usepackage{enumitem}

\usepackage{comment}

\usepackage{xspace,mfirstuc,tabulary}

\newif\iftaclinstructions
\taclinstructionsfalse 
\iftaclinstructions
\renewcommand{\confidential}{}
\renewcommand{\anonsubtext}{(No author info supplied here, for consistency with
TACL-submission anonymization requirements)}
\newcommand{\instr}
\fi

\iftaclpubformat 

\else

\fi

\title{Demarked: A Strategy for Enhanced Abusive Speech Moderation through Counterspeech, Detoxification, and Message Management}

\author{Seid Muhie Yimam \\
  University of Hamburg  \\\And
  Daryna Dementieva \\
  Technical University of Munich \\\And
  Tim Fischer \\
  University of Hamburg \\
  \AND
  Daniil Moskovskiy \\
Skoltech \\\And
  Naquee Rizwan \\
  IIT Kharagpur \\\And
  Punyajoy Saha \\
IIT Kharagpur \\\And
  Sarthak Roy \\
 IIT Kharagpur \\
 \AND
   Martin Semmann \\
University of Hamburg \\\And
  Alexander Panchenko \\
  Skoltech \\\And
  Chris Biemann \\
University of Hamburg \\\And
  Animesh Mukherjee \\
 IIT Kharagpur \\}

\begin{document}
\maketitle
\begin{abstract}


Despite regulations imposed by nations and social media platforms, such as recent EU regulations targeting digital violence, abusive content persists as a significant challenge. Existing approaches primarily rely on binary solutions, such as outright blocking or banning, yet fail to address the complex nature of abusive speech. In this work, we propose a more comprehensive approach called \emph{Demarcation} scoring abusive speech based on four aspect---(i) severity scale; (ii) presence of a target; (iii) context scale; (iv) legal scale---and suggesting more options of actions like detoxification, counter speech generation, blocking, or, as a final measure, human intervention.
Through a thorough analysis of abusive speech regulations across diverse jurisdictions, platforms, and research papers we highlight the gap in preventing measures and advocate for tailored proactive steps to combat its multifaceted manifestations. Our work aims to inform future strategies for effectively addressing abusive speech online.
\end{abstract}

\section{Introduction}

AI continues to advance rapidly across various domains, offering diverse applications. Among these, leveraging AI for societal positive impact~\cite{DBLP:journals/corr/abs-2001-01818} is becoming an extremely important direction to explore. Specifically, in the field of NLP~\cite{jin-etal-2021-good}, one of the important societal application lies in mitigating \textit{digital violence}~\cite{kaye2019speech}. 
%
Digital violence persists as a pressing issue in online social environments, posing tangible risks to users~\cite{Barbieri2019,kara2022research}. It involves using information and communication technologies to hurt, humiliate, disturb, frighten, exclude, and victimize individuals. This often results in increased anxiety, sadness, tension, and a loss of motivation at work. It includes harmful online activities such as abusive behavior, hate speech, toxic speech, and offensive language, significantly affecting an individual's professional and social effectiveness and efficiency~\cite{Fahri2022}. While this domain encompasses diverse forms of digital abuse---stalking, unauthorized photo sharing, profile hacking, and direct threats---our focus in this study is solely on \textbf{text-based abusive content}.

Traditional automated moderation methods rely mostly on only blocking of offensive messages~\cite{macavaney2019hate,cobbe2021algorithmic} to handle abusive behaviour online. Most platform companies like Meta, X etc., have a blanket hate speech policy of their own which is either deletion of the post or suspension of the account. This policy is applied uniformly across all that is flagged as hateful on the platform. Such an approach have been shown to be ineffective in curbing abusive behavior in the long term~\cite{parker2023hate}. In contrast, a more nuanced treatment of this serious problem needs to be in place. In recent times, counterspeech~\cite{DBLP:journals/corr/abs-2203-03584} has emerged as an alternative approach to mitigate hate speech and has demonstrated efficacy in addressing such harmful discourse~\cite{kulenovic2023should}. However, the automatic generation of counterspeech remains underexplored. Also, text detoxification was introduced as an approach to prevent toxic speech~\cite{nogueira-dos-santos-etal-2018-fighting,logacheva-etal-2022-paradetox}; however this method has not yet been explored in-the-wild as well.

In this work, we aim to consolidate various proactive measures for mitigating digital violence into a unified pipeline incorporating insights from several jurisdictions and NLP research in this domain. Thus, our contributions are the following.
\begin{itemize}[leftmargin=10pt]
\itemsep-0.2em  
    \item We provide a comprehensive survey of the state of hate speech definitions and mitigation strategies from three main ``pillows'': (i)~regulations across nations; (ii)~policies of social media platforms; (iii)~NLP research papers;
    \item We perform a thorough empirical analysis across a large array of documents available in these three pillows;
    \item Based on our analysis, we propose our recommendation called ``Demarcation''--- a unified pipeline for automated several-step proactive mitigation of abusive speech that consists of text detoxification, counterspeech generation, and banning via human moderation as a final step.
\end{itemize}
\vspace{-2mm}
All the questionnaires and responses utilized for this survey have been published anonymously for this submission\footnote{\url{https://anonymous.4open.science/r/demarked-tacl-D3ED/}} .






\section{Background}

Violence is an umbrella term that refers to words or actions causing harm to an individual or a community. Digital violence is a special form which anchors on digital technologies and the harm is typically spread through electronic devices like computers, smartphones and IoT sensors. This form of violence can take place publicly on social media platforms or privately on one's personal devices as well as in alternative digital environments like the metaverse. Quite naturally, the individuals or groups who are most vulnerable in the physical world are also the most vulnerable in to online abuse and harassment. In this paper, we shall only deal with textual forms of digital abuse and its nuances.



The work by \citet{banko-etal-2020-unified} classified all types of harmful content as either \textbf{abusive} or \textbf{online harm} and offered a corresponding typology. 
The typology includes four categories of harmful content as follows.
 \textbf{Hate and harassment} -  aimed at tormenting, demeaning, or intimidating specific individuals or groups, 
 \textbf{Self-inflicted harm} - content promoting self-harm,
 \textbf{Ideological harm} - the dissemination of beliefs potentially harmful to society over time, and 
 \textbf{Exploitation} - using content to exploit others financially, sexually, or physically.
The study by \citet{lewandowska2023annotation} categorizes such harmful content as \textbf{offensive speeches}, which includes 17 categories and sub-categories like \textbf{taboo}, \textbf{insulting}, \textbf{hate speech}, \textbf{harassment}, and \textbf{toxic}. The taxonomy also encompasses aspects like \textbf{hostile}, \textbf{discredit}, \textbf{racist}, among others. 
When it comes specifically to defining \textbf{hate speech}, there is no consensus among legislators, platform operators, and researchers. Furthermore, its definition has become increasingly vague amidst recent ethical and communication challenges. In addition, according to \citet{hietanen2023towards}, the definition of hate speech is now often intertwined with \textbf{negative speech}, which encompasses expressions of discontent, resentment, and blame concerning virtually any issue. One of the most comprehensive definitions that is also widely followed in the computer science literature is the one proposed by the United Nations which runs as follows - ``any kind of communication in speech, writing or behaviour, that attacks or uses pejorative or discriminatory language with reference to a person or a group based on who they are, in other words, based on their religion, ethnicity, nationality, race, colour, descent, gender or other identity factor.''
%
To address instances of textual digital violence, conventional practice often involves the implementation of \textbf{content moderation} measures. Content moderation, both human and algorithmic, involves overseeing user-generated content to align with legal standards, community norms, and platform policies \cite{banko-etal-2020-unified,hietanen2023towards}. Algorithmic moderation, primarily aimed at \textbf{removing} or \textbf{banning} non-compliant content, boosts online safety, curbs abuse, and swiftly detects serious infractions, thus reducing the limitations of depending entirely on human moderators.

Recent studies~\cite{kulenovic2023should} advocate a more sustainable method of \textbf{counterspeech} or \textbf{counter-hate} to mitigate the negative effects of hate speech. 
Counterspeech offers an approach to combating hateful content by challenging stereotypes and misinformation through reasoned arguments, thereby supporting the principle of free speech \cite{yu-etal-2022-hate,zheng-etal-2023-makes,gupta-etal-2023-counterspeeches}.
A third strategy for mitigating hate speech is \textbf{detoxification}, which seeks to decrease the toxicity level of the text maintaining content and fluency as much as feasible \cite{nogueira-dos-santos-etal-2018-fighting,DBLP:journals/mti/DementievaMLDKS21}. 
Despite being criticized by advocates of \textbf{free speech}, detoxification is designed to create a more civil digital environment for various groups, including children, but can only be used to handle explicit toxicity~~\cite{ziems-etal-2022-inducing}.

In the remainder of this paper, we will compile various methods of mitigating abusive speech in one unified pipeline of proactive moderation. 

\section{Related work}
\vspace{-2mm}
\paragraph{Automatic abusive speech detection} Moderation is a fundamental element of social media platforms, involving various measures to limit the visibility of abusive content. These measures range from deleting and hiding posts, issuing warnings, or blocking users who fail to adhere to their policies \cite{DBLP:journals/corr/abs-2312-10269,DBLP:journals/csur/AroraNHSNDZDBBA24}. Recently, significant research efforts have been focused on gathering datasets to develop automatic hate speech classification models \cite{fortuna-etal-2020-toxic2,mathew2021hatexplain}, including for low-resource languages such as Amharic \cite{ayele-etal-2023-exploring,ayele-exploring2024}, Arabic \cite{magnossao-de-paula-etal-2022-upv,alzubi-etal-2022-aixplain}, code-mixed Hindi \cite{bohra-etal-2018-dataset,ousidhoum-etal-2019-multilingual}, and others.
\vspace{-2mm}
\paragraph{Automatic counterspeech generation} While restricting access to messages remains a popular strategy endorsed by both platform owners and government policies to combat harmful content, the method of countering hate speech is increasingly favored \cite{DBLP:journals/corr/abs-2403-00179}. This strategy is frequently advocated with the motto \emph{Countering rather than censoring}, which is generally viewed as preferable since it avoids interfering with the principle of free speech \cite{DBLP:conf/emnlp/YuZ0H23,bonaldi2024nlp}. The work by \citet{DBLP:conf/emnlp/YuZ0H23} investigates counterspeech in two distinct approaches: countering the author and countering the hate content, where the former is regarded as a less robust form of countering. 
Moreover, besides reducing online hatred, counterspeech is utilized to encourage positive transformations in online communities by facilitating dialogue among users and nurturing a sense of community \cite{buerger_why_2022,doi:10.1177/20563051211063843}. 
\vspace{-2mm}
\paragraph{Automatic text detoxification} Another line of research in abusive language processing involves message detoxification. This process is crucial for eliminating or minimizing offensive or harmful content in sentences, while preserving the original meaning as much as possible  \cite{logacheva-etal-2022-paradetox,DBLP:journals/mti/DementievaMLDKS21,tran-etal-2020-towards}. Detoxification enhances the quality of online interactions by making them more respectful and less toxic \cite{tran-etal-2020-towards}. Various models applied to detoxification can produce diverse, yet non-toxic and acceptable outcomes.
\vspace{-2mm}
\paragraph{Examples of automatic mitigation strategies in deployment} The work by \citet{DBLP:journals/osnm/ChungTTG21} developed a tool for Twitter (now X) designed to continuously monitor and respond to hateful content related to Islamophobia. The tool was used by Non-governmental organization (NGO) operators, and the counter-narrative feature has been highly praised for its potential to significantly impact the fight against online Islamophobia. This feature, previously managed manually, was tested for partial automation, which proved effective in enhancing both the volume and usability of the counter-narratives produced by NGO operators.
The study by \citet{DBLP:journals/csur/AroraNHSNDZDBBA24}, using a method somehow similar to ours, examined state-of-the-art (SOTA) research on hate speech and related platform moderation policies. The findings reveal a notable discrepancy between the focus of research and the needs of platform policies. While topics like misinformation and political propaganda receive extensive research attention, as shown by the high research-to-policy coverage ratio, critical issues such as sexual solicitation and graphic content are significantly under-researched. This mismatch underscores a gap between the types of content platforms need to moderate, and the solutions offered by current harmful content detection research.
%
%
To address this perceived disconnection, we propose a dynamic and demarcation-based moderation framework, integrating multiple intervention options tailored to specific contexts and regulatory requirements.









 
\section{Methodology}
\label{methods}

As a first step toward unifying the existing hate speech regulations, our approach incorporates two strategic elements. Initially, we consider three perspectives: country-specific regulations, social media platforms' policies, and the definitions and label descriptions mentioned in research papers on hate speech. For each of these dimensions, we developed specific \textbf{selection criteria} to obtain representative samples. Subsequently, we crafted a series of \textbf{questions} designed to analyze and gain deeper insights into each area. This dual strategy ensures a comprehensive examination of the regulatory landscape.
%
%
In the following sections, we outline the selection criteria and the rationale for our questions.

\vspace{-1mm}
\subsection{Country-specific regulations}
\label{sec:countries_regulations}

In this section, we examine the national regulations on hate speech. Hate speech can manifest in various forms and requires different regulatory approaches. Our analysis aims to identify common expressions of hatred and link these findings to the digital world.
\vspace{-2mm}
\paragraph{Selection criteria}
We came up with a selection criteria to ensure a diverse and representative sample of countries, prioritizing those we are most familiar with. The criteria included:
\begin{itemize}[leftmargin=10pt]
    \itemsep-0.2em  
    \item \textbf{Home country of the co-authors}: We initially selected the home countries of all co-authors to leverage their familiarity with the regulatory framework, ensuring a detailed and contextually rich analysis.
    \item \textbf{Geographic representation}: To ensure diversity, we included at least one country from each continent. Countries were chosen based on population to capture a wide range of regulatory approaches and perspectives.
    \item \textbf{Online presence}: We selected countries that have a widespread online presence and the nationals frequently express their opinions and viewpoints on social media platforms.
    \item \textbf{Focus on hate speech regulation}: We included countries where incidences of hate speech are commonplace to gain insights into their regulatory responses to this widespread serious issue.
\end{itemize}
\vspace{-2mm}
\paragraph{Questions}
The questions formulated for regulatory analysis were guided by specific rationales aimed at extracting key insights from each country's approach to hate speech regulation.
Each question served a distinct purpose in understanding the regulatory landscape.
\begin{itemize}[leftmargin=10pt]
    \itemsep-0.2em  
    \item \textbf{Freedom of speech}: These questions aimed to assess the extent to which hate speech is protected under freedom of speech provisions, providing insights into a country's tolerance for expression that may incite hatred or discrimination.
    \item \textbf{Hate speech definition}: Identification of hate speech definitions within regulations was deemed crucial, as it reflects the country's conceptualization and legal stance on hate speech, including distinctions between online and offline manifestations.
    \item \textbf{Punishment}: Examination of punitive measures for hate speech offenses, including monetary fines and imprisonment, provided insights into the severity of regulatory responses and the enforcement mechanisms in place.
    \item \textbf{Regulation of social media platforms}: Inquiry into the regulation of social media platforms and specific provisions addressing hate speech online shed light on the extent to which countries recognize and address hate speech as a digital phenomenon.
    \item \textbf{Preventive measures}: Evaluation of regulatory encouragement for counterspeech and message detoxification initiatives reflected a country's proactive approach to combating hate speech, indicating a nuanced understanding of the issue beyond mere censorship.
\end{itemize}
\vspace{-2mm}
In total, we selected the EU and 14 other countries across the globe based as per the above selection criteria and analyzed them using our comprehensive questionnaire. The complete questionnaire are available in Appendix~\ref{sec:app_countries_questionnaire} and in \href{https://anonymous.4open.science/r/demarked-tacl-D3ED/Survey/Country%20and%20Region%20Regulations%20-%20Questions%20and%20Answers.csv}{this anonymous link}. 
The insights derived from the countries' perspectives on hate speech are discussed in Section \ref{regresul}.

\vspace{-1mm}
\subsection{Platform policies}
\label{sec:platform_policies}

Moving closer to the digital sphere, our next step involves analyzing the policies implemented by social media platforms. Our approach was structured to provide a comprehensive understanding of platform policies, considering factors such as accessibility, content moderation practices, and preventive measures.
\vspace{-2mm}
\paragraph{Selection criteria}
Our selection criteria were designed to ensure a thorough examination of policies across globally popular social media platforms while also accounting for regional variations.
These criteria encompassed the following.
\vspace{-2mm}
\begin{itemize}[leftmargin=10pt]
    \itemsep-0.2em  
    \item \textbf{Global popularity}: Platforms were selected based on their monthly active user count, prioritizing the most widely used platforms worldwide to ensure broad coverage and relevance.
    \item  \textbf{Regional relevance}: Importance was given to the popularity of platforms within the countries mentioned in Section \ref{sec:countries_regulations}.
\end{itemize}
\vspace{-2mm}
\paragraph{Questions}
The questions designed for policy analysis were crafted with specific rationales to extract crucial insights from platform policies on hate speech regulation. Each set of questions targeted a distinct aspect of platform functionalities and their strategies for addressing hate speech. The key questions are categorized as follows.
\vspace{-2mm}
\begin{itemize}[leftmargin=10pt]
   \itemsep-0.2em  
    \item \textbf{Hate speech definition}: Identifying the platform's definition of hate speech was deemed paramount, as it forms the foundation for content moderation and enforcement actions. Understanding how platforms conceptualize hate speech informs subsequent analyses of their policy effectiveness.
    \item \textbf{Platform access \& verification}: These questions sought to understand the mechanisms for user access and verification, including age restrictions and verification processes. Understanding these aspects is crucial for discerning the demographics of platform users and the potential vulnerability of certain groups, such as children, to hate speech.
    \item \textbf{Regulation accessibility}: Inquiry into the accessibility and language of platform regulations aimed to assess the transparency and user-friendliness of policy documents. In addition, examination of policy alignment with country-specific regulations provided insights into platform compliance and adaptability to legal frameworks.
    \item \textbf{Content moderation}: These questions delved into the mechanisms and actors involved in content moderation, including user-driven moderation, automated systems, and employee-led moderation teams. Insights into content moderation practices shed light on the platform's capacity to mitigate hate speech effectively.
    \item \textbf{Preventive measures}: Evaluation of preventive measures focused on the platform's efforts to empower users in reporting hate speech, as well as initiatives aimed at promoting counterspeech and detoxification of harmful content. Understanding these measures is essential for gauging the platform's commitment to combating hate speech proactively.
    \item \textbf{Data access}: Inquiry into data access policies aimed to assess the platform's transparency and willingness to collaborate with researchers and law enforcement agencies in hate speech investigations. Access to platform data is critical for conducting comprehensive research and ensuring accountability.
\end{itemize}
\vspace{-2mm}
In total, 15 social media platforms\footnote{X, Facebook, Telegram, WhatsApp, Instagram, Reddit, VK, Odnoklassniki, TikTok, YouTube, LinkedIn, Snapchat, GAB, ShareChat, Koo} were selected based on the established selection criteria and analyzed through our detailed questionnaire. The complete set of the questionnaire are available in Appendix~\ref{sec:app_platforms_questionnaire} and  \href{https://anonymous.4open.science/r/demarked-tacl-D3ED/Survey/Social%20Media%20Platforms%20Policy%20-%20Questions%20and%20Answers.csv}{this anonymous link}.
The findings, which elucidate the platforms' approaches to hate speech, are presented in Section \ref{platformresult}.

\begin{figure*}[ht!]
    
    \centering
    \begin{subfigure}[b]{0.49\textwidth}
        \includegraphics[width=1\linewidth]{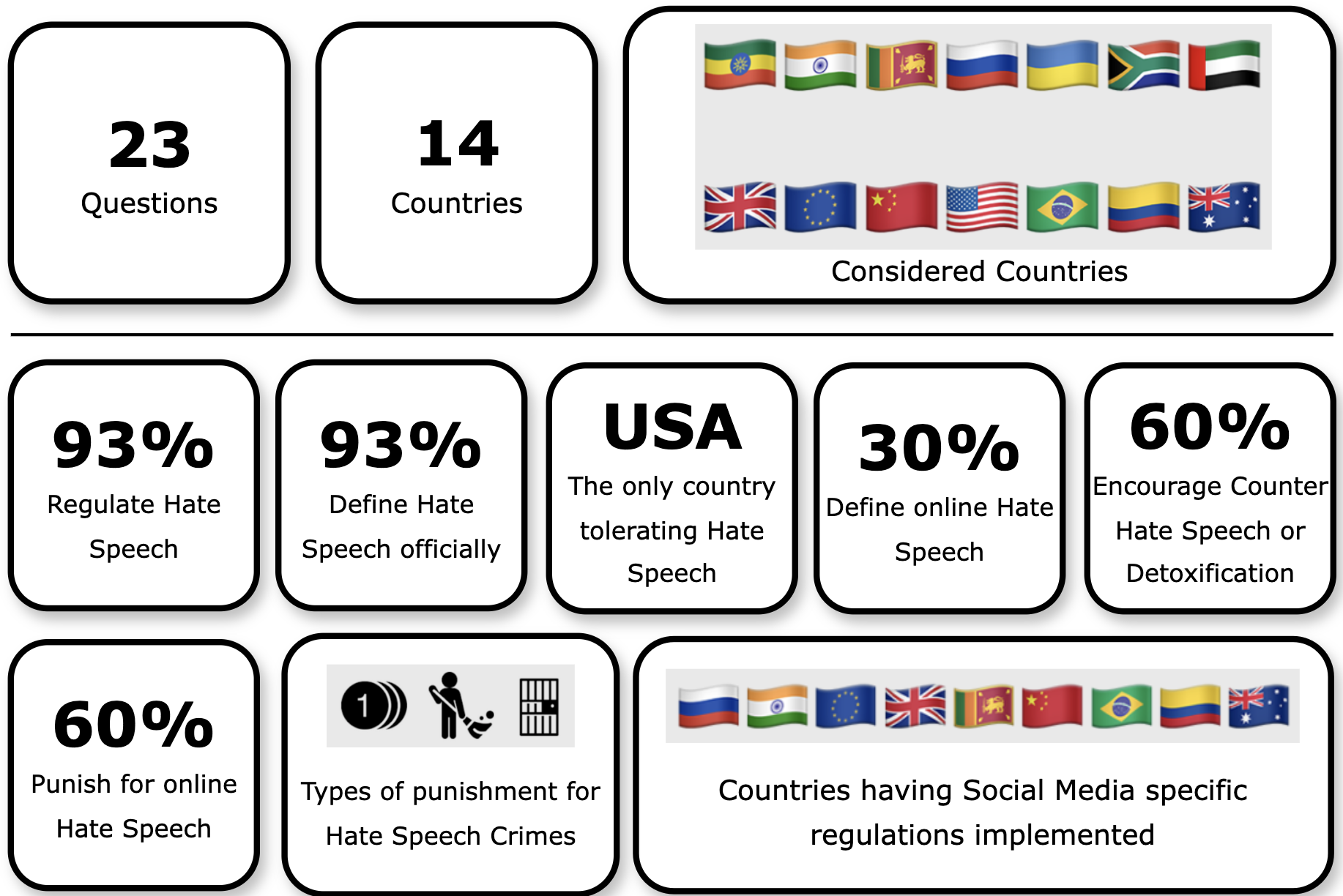}
        \caption{Regulations summary for various countries and EU.}
        \label{fig:results_countries}
   \end{subfigure}
   \hfill
   \begin{subfigure}[b]{0.47\textwidth}
        \includegraphics[width=1\linewidth]{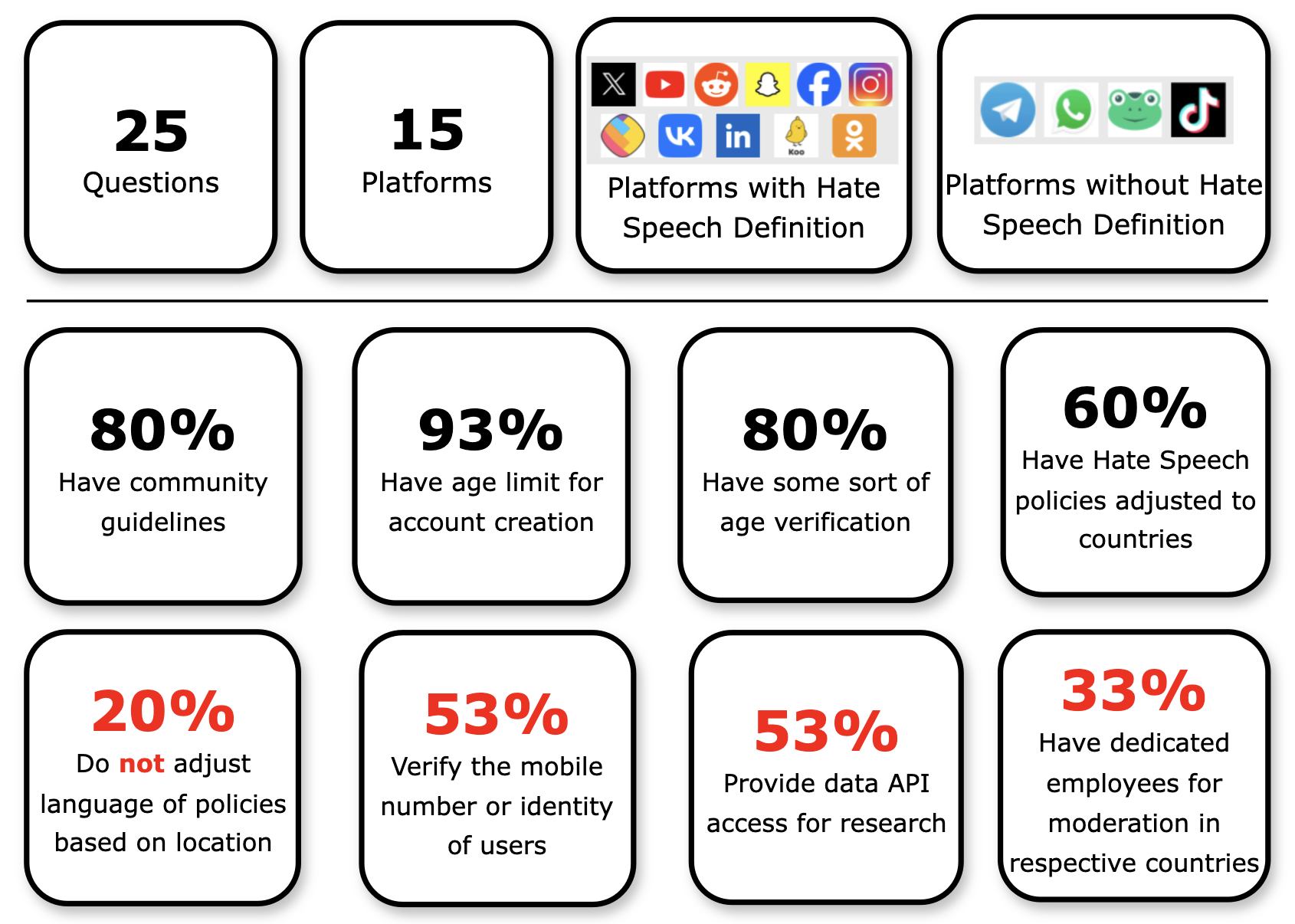}
        \caption{A summary of policies of the social platforms.}
        \label{fig:results_platforms}
   \end{subfigure}
   \caption{Exploratory results on regulations by nations and social media platforms.}
   \label{fig:results_countries_platforms}
\end{figure*}
\vspace{-1mm}
\subsection{Research datasets}
\label{sec:research_datasets}
In our third pillow, we bridge the gap with NLP research by examining the current state of automatic abusive speech detection. Our focus is on datasets designed for fine-tuning machine learning models. We ensure a thorough comprehension of the landscape across diverse languages.
\vspace{-2mm}
\paragraph{Selection criteria}
Our selection criteria were crafted to ensure the inclusion of diverse perspectives while maintaining a high standard of relevance and credibility. These criteria included the following points.
\vspace{-2mm}
\begin{itemize}[leftmargin=10pt]
    \itemsep-0.2em  
    \item \textbf{Language inclusivity}: We aimed to encompass a wide array of languages prevalent in the countries considered in Section~\ref{sec:countries_regulations}.
    \item \textbf{Citations}: We prioritized dataset papers that have significantly influenced the academic community, as indicated by their citation metrics. For low-resource languages, we included the majority or all of the available datasets to ensure comprehensive representation in our analysis.
    \item \textbf{Publication venue}: Preference was given to papers published in esteemed NLP venues such as ACL Anthology, AAAI, LREC, COLING or WOAH, ensuring a standard of quality and rigor in the selected dataset papers.
    \item  \textbf{Cross-verification}: To further bolster the credibility of our selection, we cross-checked our choices with established repositories such as hatespeechdatasets.com, thus validating the inclusion of well-established datasets.
\end{itemize}
\vspace{-2mm}
\paragraph{Questions}
The formulation of specific questions served as a scaffold for our analysis, enabling a detailed examination of the dataset papers. Each question was designed to extract key insights essential for a comprehensive understanding of hate speech datasets. 
The rationales guiding our questions were as follows:
\vspace{-2mm}
\begin{itemize}[leftmargin=10pt]
\itemsep-0.2em  
    \item \textbf{Hate speech definition}: Given the complex nature of hate speech, exploring how researchers conceptualize and define it presents valuable opportunities for deeper analysis.
    \item \textbf{Annotation process}: Investigating the annotation process sheds light on the methodologies employed, including the existence of guidelines, pilot annotations, and quality control measures, which are crucial for evaluating the quality and reliability of the dataset.
    \item \textbf{Labels}: Investigating the labels used for annotation and their descriptions provided insights into the granularity and depth of the dataset's understanding of hate speech nuances.
    \item \textbf{Annotator demographics}: Exploring the demographics of annotators, encompassing factors such as age, gender, religion, and race, facilitated an assessment of dataset inclusivity and annotator suitability.
    \item \textbf{Dataset material}: Querying aspects such as data source, modality, size, and availability is vital for understanding the dataset's scope and applicability in hate speech research.
\end{itemize}
\vspace{-2mm}
We selected 38 dataset papers spanning 20 languages based on our criteria and analyzed them using our comprehensive questionnaire. The complete questionnaire is available in Appendix \ref{sec:app_dataset_questionnaire} and in \href{https://anonymous.4open.science/r/demarked-tacl-D3ED/Survey/Research%20Datasets%20-%20Questions%20and%20Answers.csv}{this anonymous link}. 
The results from this analysis are presented in Section~\ref{sec:research_datasets_results}.








\section{Results and analysis}
In this section, we will discuss the outcomes of our investigation across three key areas aimed at mitigating hate speech: country regulations, platform policies, and research datasets.
\vspace{-1mm}
\subsection{Regulation results} 
\label{regresul}

As stated earlier, we select 14 countries from all over the world in order to have a comprehensive picture of how hate speech and related issues are regulated on a governmental level. We have at least one country from each continent and for the European region we analyzed the regulations established by the European Union which all the member states need to abide by. The results from our investigation are summarized in Figure~\ref{fig:results_countries_platforms}(a).

\vspace{-2mm}
\paragraph{Relevance of the regulations}

First of all, we note that all of the countries considered regulate hate speech in one way or another. Moreover, the absolute majority of the regulations have been updated no earlier than \textit{four} years ago, keeping the nations up-do-date with the current hate speech challenges.
\vspace{-2mm}
\paragraph{Definition of hate speech}
Despite the widespread recognition of the need to address hate speech at the a governmental level, there is no single, universally accepted definition of what constitutes hate speech. Different countries have developed their own definitions, which can lead to inconsistencies and challenges in addressing hate speech across borders. This variation in definitions highlights the need for international cooperation and dialogue to develop a shared understanding of hate speech and its consequences.
\vspace{-2mm}
\paragraph{Online hate speech regulations}

While most of the countries have laws regulating hate speech, only 60\% have specific definitions related to online hate speech. Countries such as the USA, Russia, and Ukraine do not independently address online hate speech at the legislative level, while hate speech is protected under freedom of speech in the USA.

\vspace{-2mm}
\paragraph{Punishments for hate speech}

Most countries have a tiered approach to punishing hate speech, with penalties ranging from fines and community service to imprisonment. While imprisonment is a possible consequence, the length of imprisonment is generally relatively short, with only a few countries imposing sentences exceeding 5 years.
Moreover, regulations in some countries stipulate harsher penalties for repeat offenses linked to hate crimes.
\vspace{-2mm}
\paragraph{Methods to pro-actively mitigate hate speech} At both national and regional levels, specific laws addressing counterspeech and detoxification are lacking. However, many countries have emphasized the creation of a safe environment through proactive methods, which appears to be a first positive move in this direction.

\vspace{-1mm}
\subsection{Platform results} 
\label{platformresult}
As stated earlier, we selected 15 social media platforms with the highest popularity measured in terms of monthly active users. We strategically formulated 25 questions to study the community guidelines provided by the respective platforms in terms of offensive content and their mitigation strategies. The questions can be grouped into 5 categories namely \textit{platform access and verification}, \textit{regulations}, \textit{content moderation}, \textit{preventive measures} and \textit{miscellaneous}. The overall results from our investigation are summarized in Figure~\ref{fig:results_countries_platforms}(b).
\vspace{-2mm}
\paragraph{Platform access and verification}

The majority of the platforms has an age limit for account creation and some sort of parental control. Only 3 out of 15 platforms we studied  -- Facebook, Instagram and YouTube --  apply age verification methods. A mandatory phone number or any other sort of ID verification is present in only 8 of the 15 platforms we studied. None of the platforms allow for the creation of completely anonymous accounts, but 10 platforms allow for the creation of pseudonomous accounts, i.e., an account that uses a fictitious name or alias to protect the user's digital identity.
\vspace{-2mm}
\paragraph{Regulations}

All the platforms except GAB have made their regulations or community guidelines accessible from the home page. X, Telegram and GAB do not adjust the language of the regulations automatically according to user's geographical location. Only 9 platforms have adjusted their regulations based on the country's regulations related to hate speech. Platforms like Telegram, WhatsApp, Tiktok and GAB do not even have a strict definition of hate speech in their regulations.
\vspace{-2mm}
\paragraph{Content moderation}

Platform users play an important role in content moderation where administrators or moderators can moderate respective groups or communities except Snapchat and TikTok. Platform employees play the most crucial role as content moderators, who act independently or on content flagged as offensive or inappropriate by users on all platforms except GAB. Content moderation is a subjective job and highly dependent on the social and cultural context of the individual and their demographics. Only a small minority of platforms -- Facebook, Instagram, TikTok, ShareChat, YouTube and Koo -- have moderators with demographic diversity. A common solution to this challenge is employing auto-moderation, which is adopted by almost all platforms except Telegram, WhatsApp and GAB.
\vspace{-2mm}
\paragraph{Preventive measures}

As the primary preventive measure all platforms have a reporting functionality where users can report a certain content which they find inappropriate. The users generally flag the reported content according to the category labels provided by the platform. Platforms like WhatsApp, VK, Odnoklassniki, TikTok, ShareChat and Koo do not provide a label for offensive or sensitive content while reporting. Counterspeech is one of the widely accepted counter measure for offensive speech although very few platforms like Facebook, VK and Odnoklassniki encourage the promotion of counterspeech. Another preventive measure includes detoxification of offensive content but none of the platforms have adopted this policy.

\vspace{-2mm}
\paragraph{Miscellaneous}

Platforms like VK, Odnoklassniki, GAB and ShareChat do not share data with law enforcement agencies for investigation of hateful or offensive speech. Detection and mitigation of hate speech  requires a high volume of data, which generally can be obtained from platforms via API requests. Some of the platforms such as Telegram, WhatsApp, VK, Odnoklassniki, Snapchat, GAB, ShareChat do not share data for research purposes. Influential personalities like public figures/organizations/media companies get their identities verified by the platforms but few platforms such as X, Facebook, WhatsApp, Instagram, TikTok, LinkedIn and Snapchat employ extra rules or regulations for these type of users.


\begin{figure}
    \centering
    \includegraphics[width=1\linewidth]{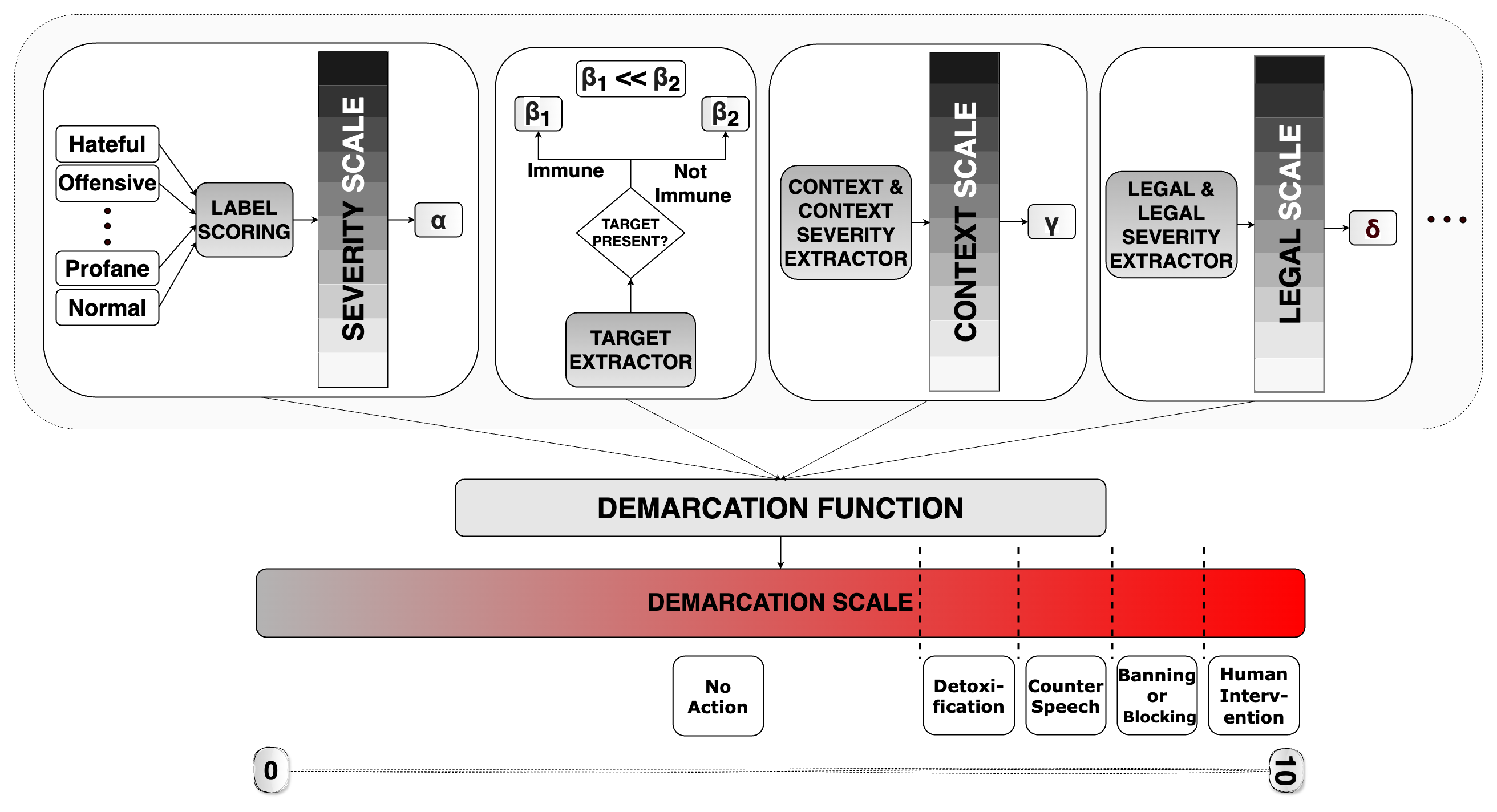}
    \caption{Demarked: the top-level architecture of our proposed proactive moderation strategy.}
    \label{fig:demarked-arch}
\end{figure}
\vspace{-1mm}
\subsection{Results based on research datasets}
\label{sec:research_datasets_results}

Our analysis of various hate speech dataset papers has yielded several key findings that provide insights into the landscape of hate speech research and dataset construction.
\vspace{-2mm}
\paragraph{Hate speech definition} 65\% of the surveyed papers present a clear definition of hate speech within their work.
We believe, especially for annotation tasks and dataset papers, conceptual clarity in understanding hate speech is highly important. Consequently, our expectation was that almost all papers would have a definition of hate speech which is unfortunately not true.
\vspace{-2mm}
\paragraph{Compliance with regulations} Our analysis reveals that only 15\% of the papers have cross-checked their definition with hate speech regulations at the national level, and only three papers referencing platform or data source-specific regulations.
This lack of alignment with regulatory frameworks highlights potential discrepancies between academic definitions and legal or platform-specific interpretations of hate speech.
\vspace{-2mm}
\paragraph{Formulation of recommendations} Only two of the 38 surveyed papers formulate recommendations on addressing hate speech or leveraging their work, datasets, or annotations for combating hate speech. We believe it is a missed opportunity for academic research to inform practical interventions and policy-making efforts in the fight against hate speech.
\vspace{-2mm}
\paragraph{Imbalance in investigated data sources} Our analysis also reveals a notable imbalance in the investigation of data sources 
 with X accounting for over 50\% of the studies, while other platforms such as YouTube, Instagram, Reddit, and WhatsApp are explored in less than 10\% of the papers. This discrepancy is significant since the over-representation of certain platforms in research does not correspond with their actual usage; for example, Facebook, with its 3 billion users, far exceeds X, which has only 611 million users.
%
While our survey is not fully representative, these findings underscore important trends and gaps in hate speech research, emphasizing the need for greater alignment with regulatory frameworks, formulation of actionable recommendations, and diversification of investigated data sources to more accurately capture the landscape of online hate speech.
Further investigation into hate speech dataset papers revealed a nuanced understanding of hate speech as a multi-faceted phenomenon. 
Through the analysis of hate speech definitions and descriptions of labels used for annotation, several key aspects emerged that can be considered for the classification of hate speech.
We outline these aspects below.
\vspace{-2mm}
\begin{itemize}[leftmargin=10pt]
\itemsep-0.2em   
    \item \textbf{Target}: Understanding the target of hate speech is essential in contextualizing its impact. Inflammatory messages directed at individuals or groups are often considered as hate speech, while undirected messages are not.
    \item \textbf{Discrimination}: Hate speech often manifests through discriminatory language targeting various characteristics such as race, sex, gender, nationality, religion, and more.
    \item \textbf{Intent of the perpetrator}: Malicious intent, ranging from mocking and causing emotional harm to issuing threats or inciting violence, is typical for hate speech. However, humorous, sarcastic or troll messages are often not considered as hate speech.
    \item \textbf{Language usage}: Hate speech can manifest in diverse linguistic forms, from threatening, dehumanizing and fear-inducing speech to overtly violent or obscene language. Again, sarcastic or humorous language is often not considered as hate speech.
    \item \textbf{Emotions of the victim/target}: Understanding the emotional impact on hate speech victims is crucial for assessing its harm, as it often induces sadness, anger, fear, and outgroup prejudice.
    \item \textbf{Frequency}: Hate speech can manifest as isolated incidents or persistent harassment, such as mobbing or bullying. Analyzing attack frequency helps gauge the severity of hate speech.
    \item \textbf{Time}: Hate speech may reference past events, current circumstances, or future actions. Especially messages that incite violent actions in the near future are dangerous. The temporal dimension should not be neglected.
    \item \textbf{Fact-checking}: Hate speech often relies on misinformation or distorted facts to perpetuate harmful narratives. Identifying disinformation can aid hate speech detection and inform the severity.
    \item \textbf{Topic and context}: Hate speech targets various topics, from political ideologies to social identities, and contextual factors must be considered in its assessment.
\end{itemize}
\vspace{-3mm}
Our analysis underscores the complexity of hate speech, highlighting the need for nuanced approaches to effectively identify, classify, and mitigate its harmful effects.
\section{Actionable recommendations}


To tackle the challenges associated with addressing abusive speech more effectively across various levels, we introduce a novel pipeline aimed at automating the process of gradually addressing harmful speech---\textit{Demarcation}. The overall schematic of our approach is presented in Figure~\ref{fig:demarked-arch}. In the rest of this section we describe the main components of our proposed pipeline together with examples of its potential deployment and implications of the future research in automatic abuse speech mitigation.
\vspace{-1mm}
\subsection{Demarcation score}
The Demarcation approach consists of several aspects that should be assessed simultaneously and combined into one final objective. The aspects that we propose to be considered are as follows.
\vspace{-2mm}
\begin{description}[leftmargin=5pt]
\itemsep-0.2em  
\item[\textbf{Severity scale} $\alpha$] First, we can detect the specific type of abusive speech and how severe it is --- if it is profane, offensive, hateful, racist, cyberbulling, etc., or just normal. Together with the label, we can estimate its severity either based on the scores from some text classification model or using additionally fine-tuned separate classifier or regressor.

\item[\textbf{Target extraction} $\beta$] 
Together with the label, it is essential to extract whether the abusive speech targets a specific victim community. The text may use rude language but remain neutral and target no one, or it could be hateful and aimed at an individual or an entire group of people.

\item[\textbf{Context scale} $\gamma$] Another important factor is a context. The conversation can be private and may consist of jokes between individuals. The situations that are crucial to handle are when the abusive speech appears in public communications. If children can be present, even profanity on the lexical level is not advisable to show. If the communication is between adults, we should mitigate hate speech, including racism, sexism, and other group-targeted aggression.

\item[\textbf{Legal scale} $\delta$] Finally, we need to take into account the platform guidelines and nationality of the perpetrator. Some expressions of hate can lead to either a ban from a social platform or even prosecutions based on the corresponding country's laws. From previous works, we can already see examples of where nations have legal regulations about online hate and platforms have special policies to cooperate with investigations. Thus, potential law violations should be specially marked and require further human intervention.
\end{description}

\vspace{-2mm}
Ultimately, we will derive the overall demarcation score for a given text instance. This computation is intended to be dynamic and may vary depending on the specific implementation. For instance, the final result could be the summation of the products of different scales:  $\alpha$,  $\beta$,  $\gamma$, and $\delta$.
\vspace{-1mm}
\subsection{Demarcation steps}
When the demarcation score $D$ is estimated, demarcation steps can be taken to reduce text toxicity and address its hatefulness. Here, we provide examples of several types of abusive speech and suggest ways to address them. However, it is important that stakeholders such as social media providers tailor these steps to the specific needs of their situation during deployment preparation.

\vspace{-2mm}
\begin{description}[leftmargin=5pt]
\itemsep-0.2em  
\item[\textbf{No counter actions}] If there is \textit{no abusive intentions} in a text sample (a text can be, for instance, even with a negative sentiment, but anyway be not abusive), then no moderation actions should be performed.

\item[\textbf{Text detoxification}] If a message contains \textit{obscene lexical elements}, as the first step of demarcation, text detoxification can be performed. This can be in the form of a suggestion to a user for transferring his/her style of profane language into civil register. Hence, this step aids in mitigating the overall toxicity of the message.

\item[\textbf{Counterspeech}] However, if a message also contains aggressiveness against some target group, then we should mitigate such \textit{hate speech} with counterspeech. This can be an even several step dialogue with explanations of why such kind of speech is inappropriate in the given context.

\item[\textbf{Blocking of a message or a user}] After proactive moderation efforts, should a user continue to ignore recommendations and repeatedly post severely abusive content that violates the platform's code of conduct or the country's rules and regulations, the published message should be blocked or deleted. The user account may also be suspended. Similarly, this measure should be enforced if the content directly incites violence.

\item[\textbf{Human moderator intervention}] At any of the previous steps, if the automatic moderation does not perform as is expected, there should be a possibility for human moderators to be notified and human intervention to occur. This process could be more refined, e.g., according to the classifier's confidence \cite{pavlopoulos-etal-2017-deeper}.

\item[\textbf{Authorities intervention}] Finally, if messages contain severe \textit{cyberbullying} or other \textit{dangerous threat speech}, high-level platform moderation should be initiated. In extreme cases, these messages should also be passed on to the appropriate department within national security forces.
\end{description}


\vspace{-2mm}
\subsection{Improved data annotation guidelines}



Within the NLP community, there are already established guidelines and data statements outlining the responsible annotation and publication of datasets~\cite{DBLP:journals/tacl/BenderF18,rogers-etal-2021-just-think,arr_responsible_checklist}. After our ``three pillows'' survey and Demarcation steps, we would like to suggest extensions of responsible datasets for specifically abusive speech annotation and publication to promote more proactive and realistic deployment of technologies based on such data.
\vspace{-2mm}
\begin{description}[leftmargin=5pt]
\itemsep-0.2em  
\item[\textbf{Country/platform alignment}] From the survey results in Section~\ref{sec:research_datasets_results}, we can observe that the foundation of the majority of the datasets were samples from popular social networks. However, only a limited number of studies have cross-referenced their definitions of abusive speech with the guidelines set by social media platforms. We recommend that future research efforts on dataset collection include definitions of abusive speech from both the social media platforms hosting the data and the countries' regulation where the target languages are predominantly spoken.
\item[\textbf{Context consideration}] Together with text instances, it would be extremely beneficial for dataset descriptions to specify the sources of these samples, whether from private conversations, interactions with bots, or public comments sections. Including additional context or preceding and following dialogue steps in these descriptions would further enhance the precision of automatic detection and moderation systems.
\item[\textbf{Moderation actions suggestion}] In the end, it is essential that authors not only provide definitions of abusive speech labels but also suggest approaches for addressing each label during moderation. Furthermore, they should dictate how to responsibly use the dataset to build models for moderating social media platforms, ensuring ethical applications and effective practices.
\end{description}


\section{Conclusion}
This work introduces \emph{Demarcation}---a novel pipeline for proactive abusive speech moderation. It is based on the demarcation score that consists of several scales: (i)~severity scale; (ii)~target extraction scale; (iii)~context scale; (iv)~legal scale. Further demarcation steps to address the impact of abusive speech should include more granular measures such as text detoxification, counterspeech, and, as ultimate mitigation actions, blocking or intervention by humans or authorities. Thus, we firmly believe that implementing a more comprehensive moderation and demarcation pipeline will enhance proactive measures against abusive speech and decrease its occurrences in the future.
We performed a thorough survey study on abusive speech regulations and mitigation strategies across three main ``pillows''---(i)~countries' regulations; (ii)~social platform policies; (iii)~NLP research papers---trying to align the real needs of healthy communications with NLP research directions. Our findings shed light on the lack of proactive moderation implementations while stressing the need of such technology at both national and social platform levels.
For this reason, in the end, we provide novel recommendations for responsible collection and publication of specifically abusive speech research data. We firmly believe that enhancing the annotation and publication standards of such datasets will contribute to the development of more accurate automatic proactive moderation models.

\vspace{-3mm}
 \section*{Limitations and Ethics statement}
While we made diligent efforts to meticulously document our research process, findings, and recommendations, it is important to acknowledge that our study initially encountered certain limitations.
%
 \textbf{1) Only text-based content}: We only took into consideration textual expression of digital violence. We acknowledge that abuse can also be extremely harmful in other modalities like images, voice recordings and videos. 
Our way of abuse mitigation do not encompass such cases.
%
\textbf{2) Only human-written content}: Our mitigation pipeline was initially tailored to address only human-authored messages and comments. However, as text generation systems become more prevalent, there is a growing influx of machine-generated content on social media platforms. It is imperative to incorporate additional measures to detect and address bots and other machine-generated texts that may pose greater risks in inciting hatred.
%
 \textbf{3) Only digital content}: Finally, we performed our studies only in the realm of digital violence. Nevertheless, digital abuse can transcend virtual platforms and manifest in real-world scenarios through various means. For this reason, we include an `authorities' intervention' step in our demarcation pipeline. 

\textbf{Ethics statement}:
We are committed to upholding freedom of speech and respect the autonomy of stakeholders in deploying moderation technologies tailored to their specific domain, context, and requirements. Our aim is to offer a broader perspective on potential automatic proactive moderation strategies, providing novel insights and recommendations.


\bibliography{custom}
\bibliographystyle{acl_natbib}


\onecolumn

\appendix

\section{Full questionnaire for ``Three Pillows''}

Below we list all the questions formulated on the country-level, social platforms hate speech regulations, and research papers.

\subsection{Countries Regulations}
\label{sec:app_countries_questionnaire}

\scriptsize {

\begin{itemize}
  \item[\textbf{Q1}] Hate speech legal definition
  \begin{itemize}
    \item[\textbf{Q1.1}] Year of the last updates on hate speech regulations (\textit{Numerical})
    \item[\textbf{Q1.2}] Are there any regulation of hate speech in the country? (\textit{Yes/No})
    \item[\textbf{Q1.3}] Is ``Hate speech'' a legal term in the law of the country? (\textit{Yes/No})
    \item[\textbf{Q1.4}] Is online hate speech defined in the regulation? (\textit{Yes/No})
    \item[\textbf{Q1.5}] Is the hate speech definition mentioned in criminal codex? (\textit{Yes/No})
    \item[\textbf{Q1.6}] Is hate speech an independent criminal offense? (\textit{Yes/No})
    \item[\textbf{Q1.7}] Is hate speech protected by freedom of speech? (\textit{Yes/No})
  \end{itemize}
  \item[\textbf{Q2}] Hate speech liability
  \begin{itemize}
    \item[\textbf{Q2.1}] Does the regulation set any kind of punishment? (\textit{Yes/No})
    \item[\textbf{Q2.2}] Is there social or community service as punishment? (\textit{Yes/No})
    \item[\textbf{Q2.3}] Is there monetary punishment? (\textit{Yes/No})
    \item[\textbf{Q2.4}] Is there imprisonment as punishment? (\textit{Yes/No})
    \item[\textbf{Q2.5}] Are there extra charge for repeated offenders? (\textit{Yes/No})
    \item[\textbf{Q2.6}] Is there special punishments for online hate speech? (\textit{Yes/No})
  \end{itemize}
  \item[\textbf{Q3}] Preventive measures
  \begin{itemize}
    \item[\textbf{Q3.1}] Do the regulations encourage counter hate speech? (\textit{Yes/No})
    \item[\textbf{Q3.2}] Do the regulations encourage message rewriting/detoxification? (\textit{Yes/No})
  \end{itemize}
    \item[\textbf{Q4}] Social media platforms
    \begin{itemize}
        \item[\textbf{Q4.1}] Are there social media platform specific regulations? (\textit{Yes/No})
        \item[\textbf{Q4.2}] Do they have social media specific regulation on hate speech? (\textit{Yes/No})
        \item[\textbf{Q4.3}] Is a time frame specified in the regulation in which a hate speech message has to be dealt with? (\textit{Yes/No})
        \item[\textbf{Q4.4}] Was the regulation updated in the last 2 years? (\textit{Yes/No})
        \item[\textbf{Q4.5}] Is the regulation of online hate speech inline with the international regulations/law? (\textit{Yes/No})
        \item[\textbf{Q4.6}] Do they have regulation of hate speech for broadcasted (TV, Radio, printed newspaper) media? (\textit{Yes/No})
    \end{itemize}
\end{itemize}

}

\label{sec:appendix}



\subsection{Social platforms regulations}
\label{sec:app_platforms_questionnaire}

\begin{itemize}
    \item[\textbf{Q1}] Hate speech legal definition
    \begin{itemize}
        \item[\textbf{Q1.1}] Company's Headquarter country (\textit{Text})
        \item[\textbf{Q1.2}] Monthly Active Users (MAU) (\textit{Numerical})
    \end{itemize}
    \item[\textbf{Q2}] Platform access \& regulations
    \begin{itemize}
        \item[\textbf{Q2.1}] Is there an age limit for account creation at the platforms? (\textit{Yes/No})
        \item[\textbf{Q2.2}] Is there content adjusted for children? (parental control?) (\textit{Yes/No})
        \item[\textbf{Q2.3}] Is there age verification? (\textit{Yes/No})
        \item[\textbf{Q2.4}] Is there phone or ID verification? (\textit{Yes/No})
        \item[\textbf{Q2.5}] Does the platform allow to create an pseudonymous account? (\textit{Yes/No})
        \item[\textbf{Q2.6}] Do they allow you to create an anonymous account? (\textit{Yes/No})
    \end{itemize}
    \item[\textbf{Q3}] Regulations accessibility
    \begin{itemize}
        \item[\textbf{Q3.1}] Are the regulations accessible from the front page? (\textit{Yes/No})
        \item[\textbf{Q3.2}] Is the regulations language automatically adjusted to the users location? (\textit{Yes/No})
        \item[\textbf{Q3.3}] Is the platform policy adjusted to the regulations of the countries? (\textit{Yes/No})
        \item[\textbf{Q3.4}] Is there a definition of hate speech?  (\textit{Yes/No})
    \end{itemize}
    \item[\textbf{Q4}] Content moderation policies
    \begin{itemize}
        \item[\textbf{Q4.1}] Are there unmoderated, private groups, channels, or chats? (\textit{Yes/No})
        \item[\textbf{Q4.2}] Is the platform moderated by users or groups? (self-moderation) (\textit{Yes/No})
        \item[\textbf{Q4.3}] Is the platform moderated by platform employees? (\textit{Yes/No})
        \item[\textbf{Q4.4}] Does the platform have a dedicated team of moderators for the specific country? (\textit{Yes/No})
        \item[\textbf{Q4.5}] Is there an auto-moderation? (pro-active moderation) (\textit{Yes/No}) 
        \item[\textbf{Q4.6}] Does the platform have community guidelines? (in addition to terms of service?) (\textit{Yes/No})
    \end{itemize}
    \item[\textbf{Q5}] Preventive measures
    \begin{itemize}
        \item[\textbf{Q5.1}] Is there a reporting functionality? (\textit{Yes/No})
        \item[\textbf{Q5.2}] Do they label content as offensive / sensitive? (\textit{Yes/No})
        \item[\textbf{Q5.3}] Do they encourage counter-hate speech? (\textit{Yes/No})
        \item[\textbf{Q5.4}] Do they have message detoxification functionality? (\textit{Yes/No})
    \end{itemize}
    \item[\textbf{Q6}] Other questions
    \begin{itemize}
    \item[\textbf{Q6.1}] Is it possible to create a group without administrator approval? (\textit{Yes/No})
    \item[\textbf{Q6.2}] Can you request data from the platform for hate speech case investigation? (usually called "Law Enforcement") (\textit{Yes/No})
    \item[\textbf{Q6.3}] Is the data accessible through API for Research? (\textit{Yes/No})
    \item[\textbf{Q6.4}] Verification of public persons / organizations / media companies? (\textit{Yes/No})
    \item[\textbf{Q6.5}] Extra rules for verified organizations, etc. (\textit{Yes/No})
    \end{itemize}
\end{itemize}



\subsection{NLP Datasets Papers}
\label{sec:app_dataset_questionnaire}

\begin{itemize}
    \item[\textbf{Q1}] Hate speech definition and alignment
        \begin{itemize}
            \item[\textbf{Q1.1}] Is there a definition of Hate speech mentioned? \textit{(Yes/No)}
            \item[\textbf{Q1.2}] What is the percentage of hateful samples? \textit{(Unknown/Not Applicable/Percentage)}
            \item[\textbf{Q1.3}] Does the paper mention alignment with countries' regulations of corresponding languages? \textit{(Yes/No)}
            \item[\textbf{Q1.4}] Does the paper mention alignment with corresponding data source's (platform) hate speech regulations? \textit{(Yes/No)}
    \end{itemize}
    \item[\textbf{Q2}] Dataset Details
        \begin{itemize}
            \item[\textbf{Q2.1}] Is the data source of the dataset mentioned? \textit{(Yes/No)}
            \item[\textbf{Q2.2}] What are the Data Source? \textit{(Unknown/List of platforms)}
            \item[\textbf{Q2.3}] What is the time period covered in the data?
\textit{(Not mentioned/Years' Range)}
            \item[\textbf{Q2.4}]  Are the target groups of the dataset specified? \textit{(Yes/No)}
            \item[\textbf{Q2.5}] Is there a clear dataset splitting strategy into train/validation/test? \textit{(Yes/No)}
            \item[\textbf{Q2.6}] Is the dataset publicly available? \textit{(Yes/On Request/No)}
            \item[\textbf{Q2.7}] What is the Dataset size (Number of Samples)? \textit{(Number)}
        \end{itemize}

    \item[\textbf{Q3}] Label details
            \begin{itemize}
                \item[\textbf{Q3.1}] Do they provide definitions for the labels? (e.g. for binary classification, is the positive class explained?) \textit{(Yes/No)}
            \item[\textbf{Q3.2}] Are the labels binary? (e.g: hate and no hate) \textit{(Yes/No)}
            \item[\textbf{Q3.3}] Are the labels fine-grained?
(e.g: Are there more annotations than just hate, no hate) \textit{(Yes/No)}
            \item[\textbf{Q3.4}] List out all the labels. \textit{(List of labels)}
            \item[\textbf{Q3.5}] Does the paper mention recommendations on how the labeled data should be used? \textit{(Yes/No)}
            \end{itemize}
    \item[\textbf{Q4}] Annotation Details
        \begin{itemize}
            \item[\textbf{Q4.1}] Do they mention the annotation tool? \textit{(Yes/No)}
            \item[\textbf{Q4.2}] What was the annotation platform? 
 \textit{(Unknown/Custom/None/System Name e.g. Toloka)}
            \item[\textbf{Q4.3}] Is the annotation conducted using crowd-sourcing? \textit{(Yes/No)}
            \item[\textbf{Q4.4}] Do they mention a pilot annotation? \textit{(Yes/No)}
            \item[\textbf{Q4.5}] Is there an annotation guideline? \textit{(Yes/No)}
            \item[\textbf{Q4.6}] Is the annotation guideline published? \textit{(Yes/No)}
            \item[\textbf{Q4.7}] What are the number of annotators per sample? 
 \textit{(Unknown/Number)}
            \item[\textbf{Q4.8}] Are there atleast 3 or more annotators? \textit{(Yes/No/Unknown)}
            \item[\textbf{Q4.9}] Do they report annotation agreement? \textit{(No/Number+Metric)}
        \end{itemize}
    \item[\textbf{Q5}] Annotator details
        \begin{itemize}
            \item[\textbf{Q5.1}] Is the payment or reward mentioned for the annotators? \textit{(Yes/No)}
            \item[\textbf{Q5.2}] Is the age of the annotators specified? \textit{(Yes/No)}
            \item[\textbf{Q5.3}] Is the gender of the annotators specified? \textit{(Yes/No)}
            \item[\textbf{Q5.4}] Is the religion of the annotators specified? \textit{(Yes/No)}
            \item[\textbf{Q5.5}] Is the race of the annotators specified? \textit{(Yes/No)}
            \item[\textbf{Q5.6}] Is the education of the annotators specified? \textit{(Yes/No)}
            \item[\textbf{Q5.7}] Is the language proficiency of the annotators specified? \textit{(Yes/No)}
            \item[\textbf{Q5.8}] Were the annotators representative of the target groups? \textit{(Yes/No)}
            \item[\textbf{Q5.9}] Do they cover therapy for the annotators? \textit{(Yes/No)}
        \end{itemize}
\end{itemize}

\newpage

\begin{table}[!ht]
    \scriptsize
    \setlength{\tabcolsep}{1pt}
    \begin{tabular}{c|cccccccccccccc}
        \toprule
        \textbf{Questions} & \textbf{Ethiopia} & \textbf{India} & \textbf{Sri Lanka} & \textbf{Russia} & \textbf{Ukraine} & \textbf{South Africa} & \textbf{UAE} & \textbf{UK} & \textbf{EU} & \textbf{China} & \textbf{USA} & \textbf{Brazil} & \textbf{Colombia} & \textbf{Australia} \\ \midrule
        \multicolumn{15}{c}{\textbf{Hate speech legal definition}} \\ \midrule
        \textbf{Q1} & 2020 & 2022 & 2024 & 2022 & 2022 & 2018 & 2023 & 2023 & 2024 & 2022 & NA & 2023 & 2019 & 2024 \\ 
        \textbf{Q1.1} & Yes & Yes & Yes & Yes & Yes & Yes & Yes & Yes & Yes & Yes & No & Yes & Yes & Yes \\ 
        \textbf{Q1.2} & No & Yes & No & No & No & Yes & Yes & No & No & No & No & Yes &  & No \\ 
        \textbf{Q1.3} & Yes & Yes & No & Yes & Yes & Yes & Yes & Yes & Yes & Yes & No & Yes & Yes & Yes \\ 
        \textbf{Q1.4} & Yes & Yes & No & No & No & Yes & No & No & Yes & No & No & No & Yes & Yes \\ 
        \textbf{Q1.5} & Yes & Yes & No & Yes & Yes & Yes & Yes & No & Yes & Yes & No & Yes & Yes & Yes \\ 
        \textbf{Q1.6} & Yes & No & No & No & No & Yes & Yes & No & No & No & No & Yes & Yes & No \\ 
        \textbf{Q1.7} & No & No & No & No & No & No & No & No & No & No & Yes & No & No & No \\ \midrule
        \multicolumn{15}{c}{\textbf{Hate speech liability}} \\ \midrule
        \textbf{Q2.1} & Yes & Yes & Yes & Yes & Yes & Yes & Yes & Yes & Yes, but & Yes & No & Yes & Yes & Yes \\ 
        \textbf{Q2.2} & Yes & No & No & Yes & No & Yes & No & No & Varies & ? & No & Yes & No & No \\ 
        \textbf{Q2.3} & Yes & Yes & No & Yes & Yes & Yes & Yes & Yes & Varies & Yes & No & Yes & Yes & Yes \\ 
        \textbf{Q2.4} & Yes & Yes & Yes & Yes & Yes & Yes & Yes & Yes & Varies & Yes & No & Yes & Yes & Yes \\ 
        \textbf{Q2.5} & No & NA & Yes & No & No & NA & NA & ? & Varies & ? & No & Yes & ? & ? \\ 
        \textbf{Q2.6} & ? & No & No & No & No & No & No & No & Varies & ? & No & No & No & ? \\ \midrule
        \multicolumn{15}{c}{\textbf{Preventive measures}} \\ \midrule
        \textbf{Q3.1} & Yes & No & No & No & No & No & No & No & No & No & Yes & Yes & No & No \\ 
        \textbf{Q3.2} & NA & No & No & No & NA & No & No & No & No & ? & Yes & No & No & No \\ \midrule
        \multicolumn{15}{c}{\textbf{Social media platforms}} \\ \midrule 
        \textbf{Q4.1} & No & Yes & Yes & Yes, partially & No & No & No & Yes & Yes & Yes & No & Yes & Yes & Yes \\ 
        \textbf{Q4.2} & Yes & Yes & No & Yes & No & No & No & Yes & Yes & Yes & No & No & Yes & Yes \\ 
        \textbf{Q4.3} & Yes & Yes & No & NA & NA & NA & No & No & Yes & Yes & No & No & Yes & Yes \\ 
        \textbf{Q4.4} & No & Yes & Yes & Yes & Yes & No & Yes & Yes & Yes & No & Yes & No & No (from 2021) \\ 
        \textbf{Q4.5} & Yes & Yes & Yes & Yes & Yes & Yes & No & ? & Yes & No & No & No & Yes & Yes \\ 
        \textbf{Q4.6} & Yes & Yes & No & Yes & Yes & No & Yes & Yes & Yes & No & Yes & No & Yes & Yes \\ \bottomrule
    \end{tabular}
    \caption{The countries regulations exploration results in terms of hate speech or digital violence regulations. We took into account countries and unions trying to cover all the continents.}
\end{table}

\begin{table*}[h!]
\centering
\footnotesize
\begin{tabular}{c|c}
\toprule
\textbf{Year of publication} & \textbf{Dataset research papers}                                                                                                                                    \\ \midrule
2017 \textit{(2)}                 & \begin{tabular}[c]{@{}c@{}}~\cite{Davidson_Warmsley_Macy_Weber_2017, Fabio_Del_Vigna_2017}\end{tabular}       \\ \midrule
2018 \textit{(7)}                & \begin{tabular}[c]{@{}c@{}}~\cite{8508247, Founta_Djouvas_Chatzakou_Leontiadis_Blackburn_Stringhini_Vakali_Sirivianos_Kourtellis_2018, bohra-etal-2018-dataset, mathur-etal-2018-detecting}\\ ~\cite{sanguinetti-etal-2018-italian, sprugnoli-etal-2018-creating, MEX-A3T-Carmona-2018}\end{tabular}                                                                                                                                       \\ \midrule
2019 \textit{(9)}                & \begin{tabular}[c]{@{}c@{}}~\cite{mulki-etal-2019-l, 10.1007/978-3-030-32959-4_18, chiril-etal-2019-multilingual}\\ ~\cite{ousidhoum-etal-2019-multilingual, 10.1145/3368567.3368584, corazza:hal-02381152}\\ ~\cite{Ptaszynski_Pieciukiewicz_Dybała_2019, fortuna-etal-2019-hierarchically, basile-etal-2019-semeval}\end{tabular}                       \\ \midrule
2020 \textit{(5)}                & \begin{tabular}[c]{@{}c@{}}~\cite{MOSSIE2020102087, sigurbergsson-derczynski-2020-offensive}\\ ~\cite{bhardwaj2020hostility, rizwan-etal-2020-hate, zueva-etal-2020-reducing}\end{tabular}                       \\ \midrule
2021 \textit{(6)}                & \begin{tabular}[c]{@{}c@{}}~\cite{9564230, 10.1007/978-981-16-0586-4_37, 11356/1462}\\ ~\cite{burtenshaw-kestemont-2021-dutch, mathew2021hatexplain, NEURIPS_DATASETS_AND_BENCHMARKS2021_c9e1074f}\end{tabular}                       \\ \midrule
2022 \textit{(8)}                & \begin{tabular}[c]{@{}c@{}}~\cite{nurce2022detecting, 10.1007/978-3-030-93709-6_41, 9971189, jeong2022kold}\\ ~\cite{das-etal-2022-hate-speech, JIANG2022100182, shekhar-etal-2022-coral, demus-etal-2022-comprehensive}\end{tabular}                       \\ \midrule
2023 \textit{(1)}                & \begin{tabular}[c]{@{}c@{}}~\cite{10076443}\end{tabular} \\ \bottomrule
\end{tabular}
\caption{The dataset research papers explored  arranged in ascending chronological order. Number in brackets denote the number of explored dataset papers published in the corresponding year.}
\label{tab:datasetCitation}
\end{table*}

\end{document}